\documentclass[letterpaper]{article} % DO NOT CHANGE THIS
\usepackage{aaai2026}  % DO NOT CHANGE THIS
\usepackage{times}  % DO NOT CHANGE THIS
\usepackage{helvet}  % DO NOT CHANGE THIS
\usepackage{courier}  % DO NOT CHANGE THIS
\usepackage[hyphens]{url}  % DO NOT CHANGE THIS
\usepackage{graphicx} % DO NOT CHANGE THIS
\usepackage{natbib}  % DO NOT CHANGE THIS AND DO NOT ADD ANY OPTIONS TO IT
\usepackage{caption} % DO NOT CHANGE THIS AND DO NOT ADD ANY OPTIONS TO IT
\usepackage{algorithm}
\usepackage{algorithmic}
\usepackage{siunitx}
\usepackage{multirow}
\usepackage{booktabs}

\usepackage{amsthm,amsmath,amssymb}
\usepackage[capitalize,noabbrev]{cleveref}

\usepackage{newfloat}
\usepackage{listings}
\DeclareCaptionStyle{ruled}{labelfont=normalfont,labelsep=colon,strut=off} % DO NOT CHANGE THIS
\floatstyle{ruled}
\newfloat{listing}{tb}{lst}{}
\floatname{listing}{Listing}
\title{Playmate2: Training-Free Multi-Character Audio-Driven Animation\\ via Diffusion Transformer with Reward Feedback}
\author{
    Xingpei Ma\equalcontrib,
    Shenneng Huang\equalcontrib,
    Jiaran Cai\equalcontrib\thanks{Project lead \& Corresponding Author.},
    Yuansheng Guan\equalcontrib,
    Shen Zheng\equalcontrib,
    Hanfeng Zhao,
    Qiang Zhang,
    Shunsi Zhang
}
\affiliations{
    Guangzhou Quwan Network Technology \\
    \{maxingpei, huangshenneng, caijiaran, guanyuansheng, zhengshen, zhaohanfeng, zhangqiang, zhangshunsi\}@52tt.com \\[0.4em]
    \centering
    \includegraphics[width=\linewidth]{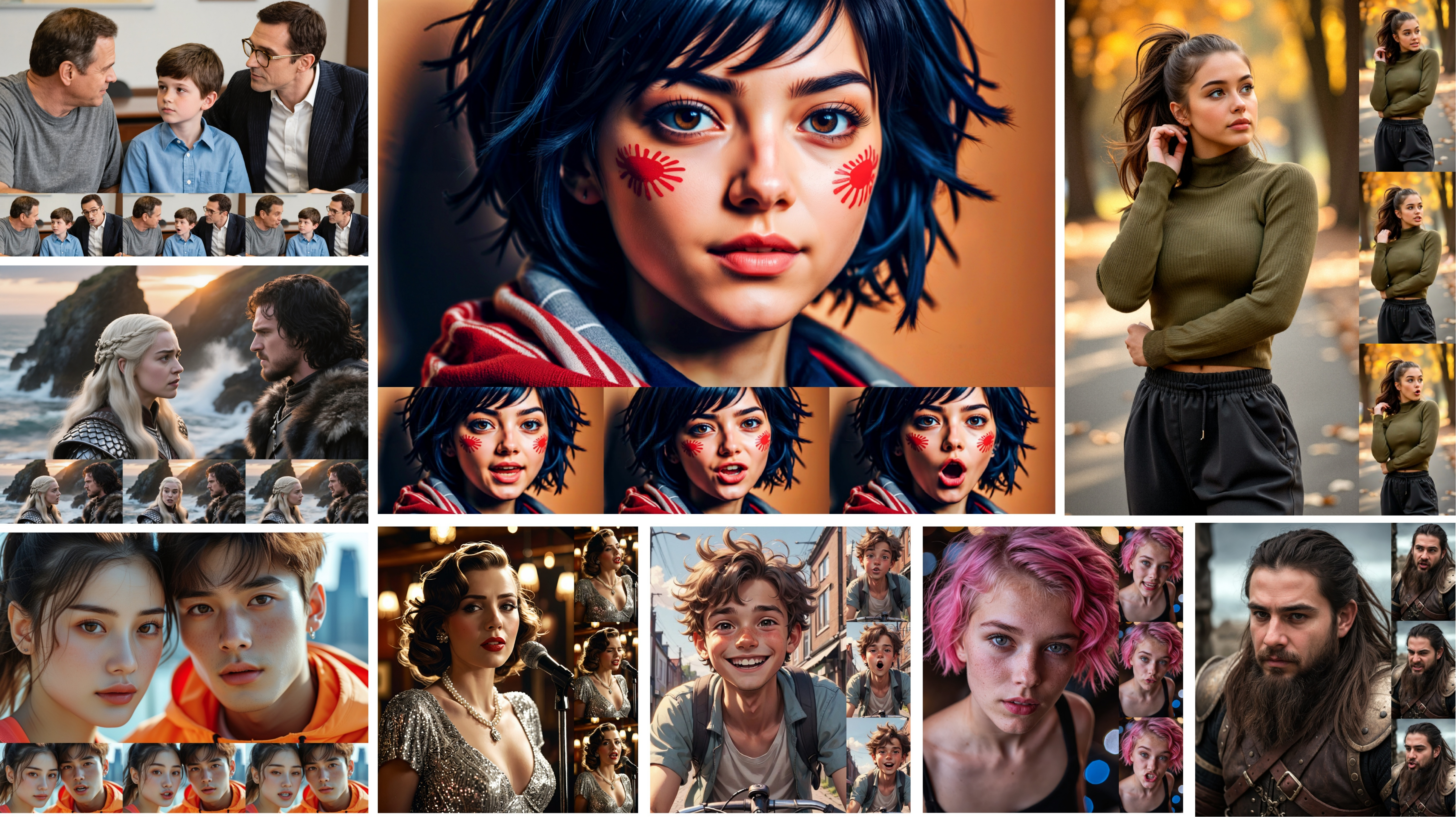}
    \begin{flushleft}
    {Figure 1: We present a novel DiT-based framework for generating high-quality, audio-driven human videos that effectively tackles key challenges related to temporal coherence in long sequences and multi-character animations. To the best of our knowledge, this is the first training-free approach capable of enabling audio-driven animation for three or more characters without requiring additional data or model modifications.}
    \end{flushleft}
    \vskip -0.3in
}

\usepackage{bibentry}
\begin{document}

\maketitle
% \twocolumn[{
% \renewcommand\twocolumn[1][]{#1}
% \begin{center}
%     \maketitle
%     \centering
%     \includegraphics[width=\linewidth]{./figures/showcase.pdf}
%     \begin{flushleft}
%     {Figure 1: We present a novel DiT-based framework for generating high-quality, audio-driven human videos that effectively tackles key challenges related to temporal coherence in long sequences and multi-character animations. To the best of our knowledge, this is the first training-free approach capable of enabling audio-driven animation for three or more characters without requiring additional data or model modifications.}
%     \end{flushleft}
% \end{center}
% }]

\setcounter{figure}{1}

\begin{abstract}
Recent advances in diffusion models have significantly improved audio-driven human video generation, surpassing traditional methods in both quality and controllability. However, existing approaches still face challenges in lip-sync accuracy, temporal coherence for long video generation, and multi-character animation. In this work, we propose a diffusion transformer (DiT)-based framework for generating lifelike talking videos of arbitrary length, and introduce a training-free method for multi-character audio-driven animation. First, we employ a LoRA-based training strategy combined with a position shift inference approach, which enables efficient long video generation while preserving the capabilities of the foundation model. Moreover, we combine partial parameter updates with reward feedback to enhance both lip synchronization and natural body motion. Finally, we propose a training-free approach, Mask Classifier-Free Guidance (Mask-CFG), for multi-character animation, which requires no specialized datasets or model modifications and supports audio-driven animation for three or more characters. Experimental results demonstrate that our method outperforms existing state-of-the-art approaches, achieving high-quality, temporally coherent, and multi-character audio-driven video generation in a simple, efficient, and cost-effective manner.
\end{abstract}

\begin{links}
    \link{Project Page}{https://playmate111.github.io/Playmate2/}
\end{links}

\section{Introduction}
Benefiting from large-scale pre-training and advanced architectural designs, diffusion models~\cite{rombach2022high,esser2024scaling,liu2024sora,yang2024cogvideox,kong2024hunyuanvideo,wan2025wan} have achieved significant advances in image and video synthesis, outperforming traditional generative adversarial networks (GANs)~\cite{goodfellow2020generative} in both visual quality and temporal coherence. These advancements have greatly improved the generation of audio-driven human videos~\cite{xue2024human}, making this capability a cornerstone of digital human research. Audio-driven human animation has broad applications in digital entertainment, film and gaming production, virtual reality, and digital storytelling. 

Audio-driven human animation~\cite{jiang2024audio} synthesizes realistic character videos with synchronized lip movements and natural body gestures from speech and auxiliary inputs. Recent advances in diffusion models have spurred their use in this domain, leading to two main categories: portrait animation and human animation. The first focuses on synthesizing facial expressions solely from audio signals, with little attention given to background dynamics~\cite{tian2024emo,xu2024hallo,xu2024vasa,ji2024sonic,ma2023dreamtalk,chen2024echomimic,cui2025hallo3}. Such a restricted approach frequently compromises the realism of generated videos in complex scenes, leading to results that do not satisfy the demands of high-quality applications. The second employs video diffusion models to overcome the aforementioned spatial constraints, thereby achieving full-body animation generation~\cite{lin2025omnihuman,fei2025skyreels,wang2025fantasytalking,chen2025hunyuanvideo,kong2025let,cui2025hallo4}. Despite progress, several challenges persist: 1) Existing methods often struggle to maintain accurate lip-sync while generating natural body movements; 2) In long video synthesis, current solutions often result in jittery motions and abrupt transitions, failing to preserve temporal coherence; 3) Most existing techniques are unable to animate scenes involving multiple characters using audio input; although some works achieve multi-character animation by constructing multi-speaker datasets and introducing significant modifications to the model architecture, such strategies are often resource-intensive and not scalable.

To address these challenges, leveraging the large-scale video diffusion model Wan2.1~\cite{wan2025wan}, we propose a diffusion transformer (DiT)-based~\cite{peebles2023scalable} framework for audio-driven facial and human video generation, aiming to enhance video quality and enable cost-effective multi-character animation. First, we adopt a LoRA-based~\cite{hu2022lora} training strategy to preserve the capabilities of the foundation model while enabling long video generation. Next, we explore a training strategy that combines partial parameter updates with reward feedback, producing videos with accurate lip synchronization and natural body motions. Finally, inspired by Classifier-Free Guidance (CFG)~\cite{ho2022classifier}, we introduce a training-free approach, Mask Classifier-Free Guidance (Mask-CFG), to tackle the challenge of multi-character animation. This method does not require constructing specialized datasets or modifying the model architecture; instead, it achieves multi-character control through simple adjustments during inference, making it both efficient and cost-effective. 

To the best of our knowledge, this is the first training-free approach capable of enabling audio-driven animation for three or more characters. In summary, our contributions are as follows: 

\begin{itemize}
    \item We propose a DiT-based framework for audio-driven human animation, combined with a LoRA-based training strategy for long video generation. 
    \item We investigate a training strategy combining partial parameter updates with reward feedback to improve lip-sync accuracy while maintaining natural and adaptive body movements.
    \item We introduce a training-free method (Mask-CFG) to support multi-character animation, which is both efficient and cost-effective.
\end{itemize}

\section{Related Work}
\begin{figure*}[!t]
\centering
\includegraphics[width=0.95\textwidth]{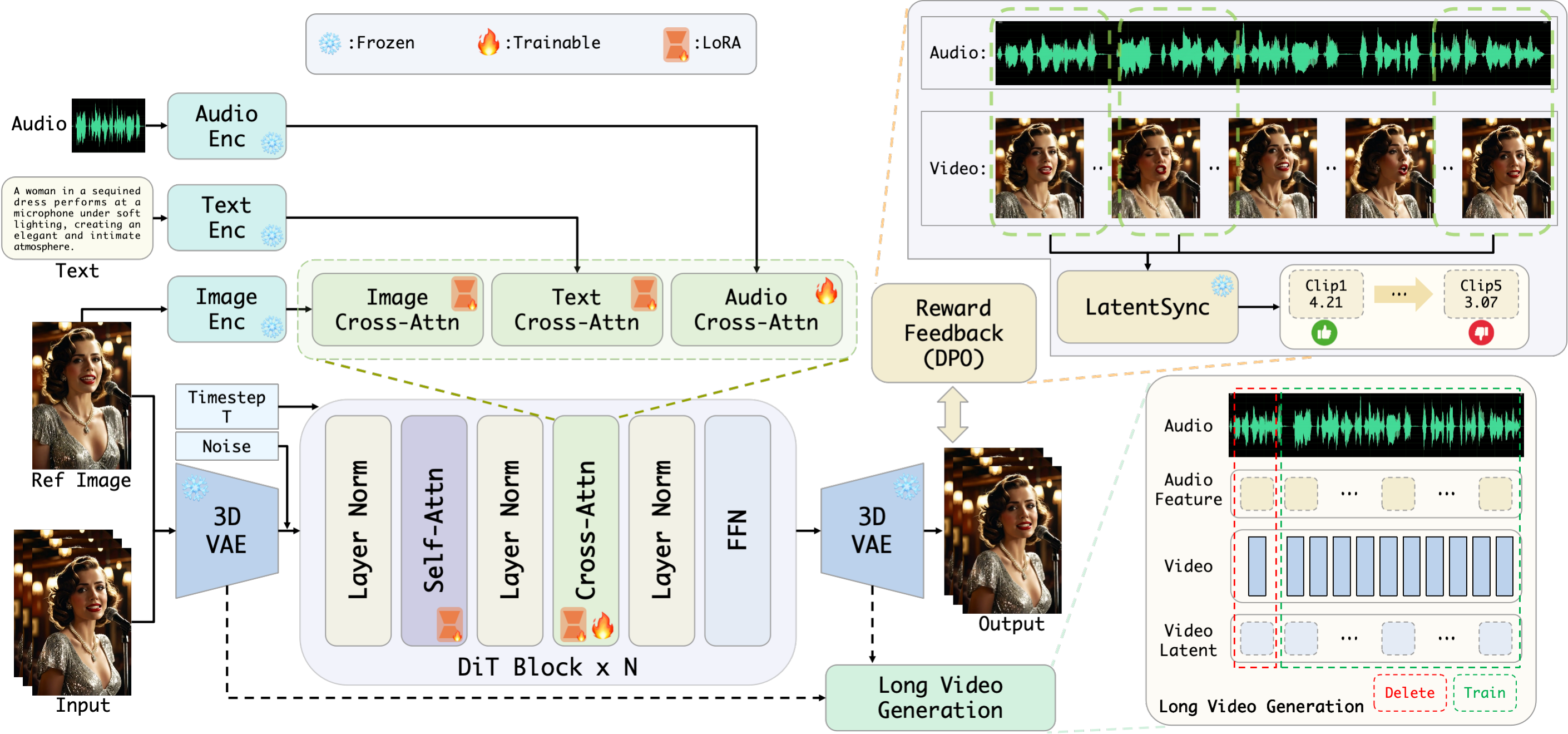}
\caption{Overview of our method. Our framework leverages a LoRA-based training strategy and position shift inference to generate long, temporally coherent videos with consistent identity. A partial parameter update with reward feedback enhances lip synchronization and upper-body motion naturalness. Furthermore, we propose Mask-CFG, a training-free approach for multi-character animation that requires no additional data or model fine-tuning, yet supports audio-driven animation of three or more characters.}
\label{fig:framework}
\end{figure*}

\subsection{Audio-Driven Portrait Animation}
Prior work on audio-driven portrait animation has largely focused on lip-sync accuracy~\cite{prajwal2020lip,zhang2023dinet,zhang2023sadtalker,guo2021ad,wang2024expression}. Traditional approaches based on GANs, neural radiance fields (NeRF)~\cite{mildenhall2021nerf}, and 3D Gaussian Splatting~\cite{kerbl20233d} have achieved strong results, yet often fail to model the subtle relationship between prosody and facial dynamics, leading to limited expressiveness and reduced visual realism. Recently, diffusion-based methods have enabled end-to-end talking video generation. EMO~\cite{tian2024emo} improves inter-frame consistency for stable, natural synthesis. Hallo~\cite{xu2024hallo} jointly addresses lip synchronization, expression, and pose. Sonic~\cite{ji2024sonic} emphasizes global perceptual coherence for diverse motions, while DICE-Talk~\cite{tan2025disentangle} and Playmate~\cite{maplaymate} introduce emotional control for expressive portraits. Despite producing realistic outputs, these methods are primarily limited to facial animation and do not support full-body motion synthesis.

\subsection{Audio-Driven Human Animation}
To enable audio-driven human animation, recent methods leverage large-scale video diffusion models. CyberHost~\cite{lin2024cyberhost} proposes a one-stage framework with novel attention and human-prior-guided training for upper-body synthesis. Approaches like OmniHuman-1~\cite{lin2025omnihuman}, FantasyTalking~\cite{wang2025fantasytalking}, SkyReels-Audio~\cite{fei2025skyreels}, and OmniAvatar~\cite{gan2025omniavatar} build on models such as Seaweed~\cite{seawead2025seaweed}, HunyuanVideo~\cite{kong2024hunyuanvideo}, and Wan2.1 for holistic motion generation. In multi-character scenarios, HunyuanVideo-Avatar~\cite{chen2025hunyuanvideo} uses latent-space masking for localized, character-specific control, while MultiTalk~\cite{kong2025let} introduces Label Rotary Position Embedding with a multi-person dataset to resolve audio-person binding. Inspired by these advances, we base our approach on a large-scale video diffusion transformer for audio-driven human animation.

\subsection{Direct Preference Optimization}
Reinforcement Learning from Human Feedback (RLHF)~\cite{ouyang2022training} is widely used to align large language models with human preferences. This paradigm has been extended to image and video generation via reward models or preference data~\cite{zhang2024direct,wallace2024diffusion,liu2025improving}. Recently, Direct Preference Optimization (DPO)~\cite{rafailov2023direct} has gained traction in audio-driven animation. Hallo4~\cite{cui2025hallo4} proposes a DPO framework for human-centric animation, leveraging a curated human preference dataset to align generated outputs with perceptual metrics related to motion-video alignment and facial expression naturalness. EchoMimicV3~\cite{meng2025echomimicv3} further adopts an alternating Supervised Fine-Tuning (SFT) and DPO training paradigm, enabling high-quality video generation with a 1.3B-parameter model. Building on these advances, we present a more efficient framework that integrates DPO to simultaneously enhance lip synchronization accuracy and facial expression naturalness in audio-driven video generation.

\section{Methodology}
Our method generates high-quality talking videos and enables efficient multi-character animation from a single image, text prompt, and audio clip. The overall framework is illustrated in \cref{fig:framework}. Built upon the Wan2.1 video diffusion model, we propose a DiT-based architecture enhanced with a LoRA-based training strategy to support long-duration video generation (\cref{subsec:3_1}). We further introduce a partial-update training approach with reward feedback to improve visual fidelity (\cref{subsec:3_2}) and lip synchronization accuracy. Finally, we present a training-free solution, named Mask-CFG, for efficient and cost-effective multi-character audio-driven animation (\cref{subsec:3_3}).

\subsection{LoRA-based Long Video Generation}
\label{subsec:3_1}
The framework of our method is illustrated in \cref{fig:framework}, where Wan2.1 serves as the foundational model. Specifically, we employ the causal 3D Variational Autoencoder (VAE)~\cite{kingma2013auto} from Wan2.1 to compress both the reference image and the ground-truth video from pixel space to the latent space. Additionally, we use UMT5~\cite{chung2023unimax} for text encoding and CLIP~\cite{radford2021learning} for image encoding. For audio input, we utilize Wav2Vec~\cite{baevski2020wav2vec} to extract audio tokens containing rich multi-scale acoustic features, which are then injected into the DiT through cross-attention mechanisms. 

HunyuanVideo-Avatar~\cite{chen2025hunyuanvideo} uses the Time-Aware Position Shift Fusion method from Sonic~\cite{ji2024sonic} to enable long video generation. OmniAvatar~\cite{gan2025omniavatar} reuses the final latent of the current segment as the initial latent for the next, and applies reference image embedding to preserve identity and maintain frame overlap for temporal consistency. Our experiments show that these two methods fail to achieve satisfactory performance in long video generation. This issue stems from the special architecture of video diffusion models, such as Wan2.1, which are designed to support joint training on both video and image data. In particular, given an input video $V \in \mathbb{R}^{(1 + T)\times H\times W\times 3}$, where the frames of $V$ follow the $1+T$ input format, Wan2.1 divides the video into $1 + T/4$ chunks. Then, Wan-VAE compresses the spatio-temporal dimensions of these chunks to $[1 + T/4,H/8,W/8]$, while the first frame is only spatially compressed to better handle image data. This independent processing of the first frame tends to cause forgetting and drifting issues.

To address this issue, we divide the video into $T/4$ chunks and encode each chunk into a single latent representation. Subsequently, we employ the LoRA training approach, which enables the model to efficiently adapt to long video generation while maintaining high-quality output and low computational cost during training. Notably, we do not add audio cross-attention layers at this stage; instead, we apply LoRA training only to the self-attention and cross-attention modules within the Wan2.1 DiT blocks.

\subsection{Partial-update Training and DPO}
\label{subsec:3_2}
\subsubsection{Audio Cross-Attention.} After completing the first LoRA-based training stage, we obtain a diffusion transformer capable of seamless long video generation. Next, we introduce the Audio Cross-Attention module and adopt the Flow Matching~\cite{lipman2022flow} approach used in Wan2.1 to update its parameters. Specifically, we aggregate every four consecutive audio frames into a single representation to ensure temporal alignment between the audio features and the compressed video latent representation. The Audio Cross-Attention mechanism is defined as:
\begin{equation}
z'=\text{CrossAttn}(z_v,z_a)=\text{Attn}(Q_v,K_a,V_a),
\end{equation}
where $z_v \in \mathbb{R}^{b\times f\times (w\times h)\times c}$ and $z_a \in \mathbb{R}^{b\times f\times l\times c}$ denote the video and audio tokens, respectively. Here, $f$, $h$ and $w$ represent the number of frames, height, and width of the latent video representation, while $l$ denotes the sequence length of the audio tokens. $Q_v$, $K_a$, and $V_a$ are the video query, audio key, and audio value matrices, respectively. 

Finally, we use the following Flow Matching objective to update the parameters of the Audio Cross-Attention module:
\begin{equation}
{\mathcal{L}}=\mathbb{E}_{z_0,z_1,z_a,t}\left\| v_{\theta_a}\left(z_t,z_a,t;\theta_a \right)-v_t\right\|^{2},
\label{eq:l_diff}
\end{equation}
where $z_1$ denotes the latent embedding of the training sample, and $z_0$ denotes the initial noise sampled from the standard Gaussian distribution $\mathcal{N}(\mathbf{0}, \mathbf{I})$. The latent variable $z_t$ is linearly interpolated between $z_0$ and $z_1$, and its time derivative $v_t = \frac{d z_t}{d t} = z_1 - z_0$ serves as the regression target. The model predicts this velocity as $v_{\theta_a}\left(z_t,z_a,t;\theta_a \right)$, where $\theta_a$ represents the parameters of the Audio Cross-Attention module, and $z_a$ denotes the audio features used for conditioning.

\subsubsection{Reward Feedback.} To further improve lip-sync accuracy and align the model with human preferences, we introduce DPO for optimization after completing the aforementioned stages. Hallo4~\cite{cui2025hallo4} presents the first audio-driven portrait DPO dataset that captures human preferences in lip-sync and facial naturalness via annotator rankings. Unlike Hallo4, which relies on human annotators to construct the dataset, we introduce DPO in a more efficient and cost-effective manner. 

Direct Preference Optimization formulates the alignment of models with human preferences as a policy optimization task, based on pairwise preference data $\mathcal{D} = \{(x, y^w, y^l)\}$, where $y^w$ is preferred over $y^l$. As shown in \cref{fig:framework}, for each training sample, we randomly select five segments and employ LatentSync~\cite{li2024latentsync} to compute the Sync-C score for each; the highest-scoring segment is selected as $y^w$, and the lowest as $y^l$. Finally, we use the Flow-DPO loss proposed by VideoReward~\cite{liu2025improving} to train the model:
\begin{equation}
\begin{split}
\mathcal{L}_{\text{DPO}} = 
&-\mathbb{E}_{y^w,y^l,t} \biggl[ \log \sigma \biggl( -\frac{\beta_t}{2} \Bigl( 
  \| v^w - v_{\theta_a}(y_t^w, t) \|^2   \\
&- \| v^w - v_{\text{ref}}(y_t^w, t) \|^2 
- \| v^l - v_{\theta_a}(y_t^l, t) \|^2 \\
&+ \| v^l - v_{\text{ref}}(y_t^l, t) \|^2 \Bigr) \biggr) \biggr],
\end{split}
\label{eq:l_dpo}
\end{equation}
where $v_{\text{ref}}$ denotes the reference model, initialized from the previously fine-tuned diffusion model; $v^w$ and $v^l$ denote the velocity fields derived from the preferred sample $y^w$ and the dispreferred sample $y^l$, respectively. Here, $\beta_t = \beta (1-t)^2$, and the expectation is taken over $(y^w, y^l) \sim \mathcal{D}$ and $t \sim [0,1]$. The overall training loss during this stage is:
\begin{equation}
\mathcal{L}_{\text{all}} = \mathcal{L}_{\text{diff}} + \lambda \mathcal{L}_{\text{DPO}},
\end{equation}
where $\mathcal{L}_{\text{diff}}$ and $\mathcal{L}_{\text{DPO}}$ denote the losses in \cref{eq:l_diff} and \cref{eq:l_dpo}, respectively, and $\lambda$ is set to 0.1.

\subsection{Mask-CFG for Multi-Character Audio-driven Animation}
\label{subsec:3_3}
\begin{figure}[!t]
\centering
\includegraphics[width=0.95\columnwidth]{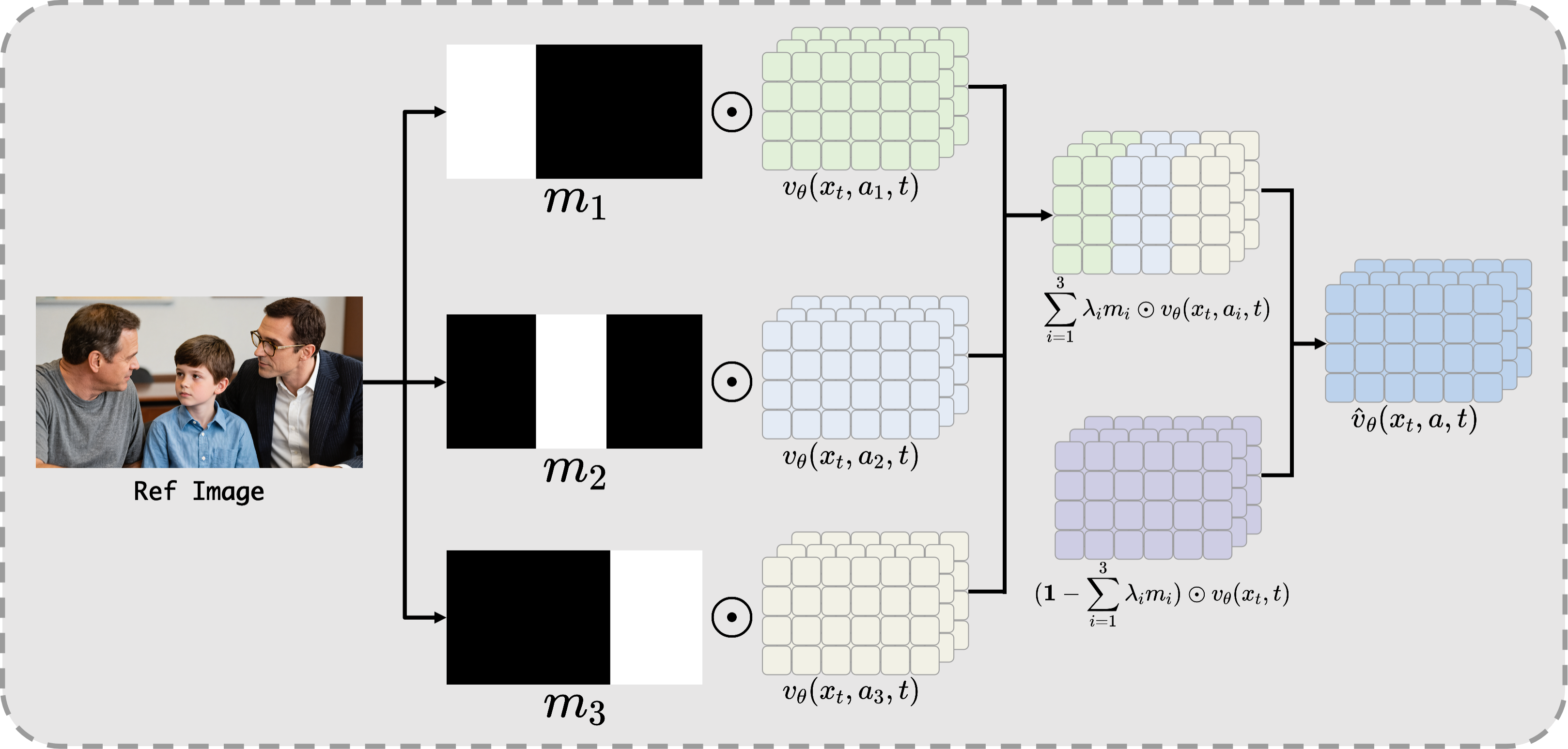}
\caption{Workflow of Mask-CFG.}
\label{fig:mask-cfg}
\end{figure}

After the training stage described above, we obtain a diffusion transformer that achieves accurate lip-sync and strong alignment with human preferences. We now introduce a training-free method to enable multi-character audio-driven video generation. Methods such as MultiTalk~\cite{kong2025let} and HunyuanVideo-Avatar achieve this capability by constructing multi-speaker datasets and modifying the cross-attention mechanism. In contrast, we enable multi-character animation by improving the classifier-free guidance (CFG) mechanism during inference—without any training or model modification—resulting in a simple, efficient, and framework-agnostic approach.
Specifically, we propose Mask-CFG, which leverages spatial masks to route audio conditions to specific characters. Given an audio condition set $A = \{ a_1, a_2, \dots, a_n \}$, the corresponding binary mask set is defined as $M = \{ m_1, m_2, \dots, m_n \}$. Here, $a_1$ is considered to be silent audio, and $m_1$ serves as the background mask. Each $m_i \in \{0,1\}^{H \times W}$ denotes a binary segmentation of the input image, and the masks are exhaustive and mutually exclusive, satisfying $\bigvee_{i=1}^{n} m_i = \mathbf{1}$, meaning their union covers the entire image region. Under classifier-free guidance, the conditional distribution $p(a_i \mid x_t)$ leads to the following formulation:
\begin{equation}
\begin{split}
p(a_i\mid x_t)&=p\left(a_i \,\middle|\, \sum_{j=1}^{n} m_j \odot x_t \right) \\
&=\frac{p(\sum_{j=1}^{n}m_j\odot x_t\mid a_i)p(a_i)}{p(\sum_{j=1}^{n}m_j\odot x_t )} \\
&=\frac{\prod_{j=1}^{n}p(m_j\odot x_t\mid a_i)p(a_i)}{\prod_{j=1}^{n}p(m_j\odot x_t)} \\
&=\frac{p(m_i\odot x_t\mid a_i)p(a_i)\prod_{j=1,j\ne i}^{n}p(m_j\odot x_t)}{p(m_i\odot x_t)\prod_{j=1,j\ne i}^{n}p(m_j\odot x_t )} \\
&=\frac{p(m_i\odot x_t,a_i)}{p(m_i\odot x_t)} \\
&=p(a_i\mid m_i\odot x_t).
\end{split}
\label{eq:cfg_1}
\end{equation}
Substituting $p(a_i\mid x_t) = p(a_i\mid m_i\odot x_t)$ into the CFG score term $\nabla_{x_t} \log p(a_i\mid x_t)$ and combining it with the standard CFG formula, we obtain:
\begin{equation}
\begin{split}
\hat{v}_{\theta }(&x_t,a,t)=\nabla_{x_t}\log p(x_t)+\lambda \nabla_{x_t}\log p(a\mid x_t) \\
&=\nabla_{x_t}\log p(x_t) + \lambda \nabla_{x_t} \log \prod_{i=1}^{n} p(a_i\mid x_t) \\
&=\nabla_{x_t}\log p(x_t) + \lambda \sum_{i=1}^{n} \nabla_{m_i \odot x_t} \log p(a_i\mid m_i \odot x_t) \\
&=\nabla_{x_t}\log p(x_t) + \lambda \sum_{i=1}^{n} m_i \odot \nabla_{x_t} \log p(a_i\mid x_t) \\
&={v}_{\theta }(x_t,t) + \sum_{i=1}^{n} \lambda_i m_i \odot [v_{\theta }(x_t,a_i,t)-v_{\theta }(x_t,t)].
\end{split}
\label{eq:cfg_2}
\end{equation}
Through the above Mask-CFG approach, we achieve multi-character audio-driven video generation in a training-free manner, with the visualization workflow shown in \cref{fig:mask-cfg}.

\section{Experiment}
\subsection{Experimental Setups}
\label{subsec:4_1}
\subsubsection{Datasets.} We collect our training data from public datasets (including AVSpeech~\cite{ephrat2018looking} and OpenHumanVid~\cite{li2025openhumanvid}) and sources we collect ourselves. To ensure high data quality, we employ tools such as Koala-36M~\cite{wang2025koala} to filter out videos with low brightness or poor aesthetic quality. Through this standardized selection process, we obtain over \num[group-separator={,}]{300000} training samples, with a total duration exceeding \num[group-separator={,}]{800} hours. To demonstrate the effectiveness of our multi-character audio-driven approach in a training-free manner, all training samples are single-person talking videos. For evaluation, we use two public datasets: CelebV-HQ~\cite{zhu2022celebv}, which features diverse scenes, and HDTF~\cite{zhang2021flow}, which provides high-resolution videos and a larger number of subjects, to assess the animation capabilities of our method.

\subsubsection{Implementation Details.} The training weights of our first LoRA stage are initialized from the pretrained Wan2.1-I2V-14B-720P model, and the training process is conducted using \num[group-separator={,}]{16} NVIDIA A100 GPUs for a total of \num[group-separator={,}]{5000} steps. Subsequently, we use \num[group-separator={,}]{32} NVIDIA A100 GPUs and conduct additional training for \num[group-separator={,}]{100000} steps to obtain the first $v_\text{ref}$. Next, we introduce DPO-based refinement training for another \num[group-separator={,}]{100000} steps, during which $v_\text{ref}$ is updated every \num[group-separator={,}]{10000} steps. The model operates at a resolution of $720 \times 1280$. We employ AdamW as the optimizer and set the learning rate to $1 \times 10^{-5}$. After completing the above training pipeline, we introduce Mask-CFG during the inference stage to enable multi-person audio-driven animation, with $\lambda$ set to \num[group-separator={,}]{5.0}.

\subsubsection{Evaluation Metrics.} We evaluate the superiority of our method using several widely adopted metrics from prior work. Specifically, we employ the Fréchet Inception Distance (FID) and Fréchet Video Distance (FVD) to assess the visual quality and diversity of the generated content. Audio-visual synchronization is measured using Sync-C and Sync-D. Furthermore, we conduct an analysis of both perceptual quality, using Image Quality Assessment (IQA), and aesthetic appeal with the Aesthetic Score Estimator (ASE). 

\begin{table*}[!t]
  \centering
  \renewcommand{\arraystretch}{1.5}
  \begin{tabular}{l|cccccc}
  \toprule
  \multirow{2}{*}{Method} & \multicolumn{6}{c}{HDTF/CelebV-HQ} \\
    % \cline{2-7}
    ~ & FID $\downarrow$ & FVD $\downarrow$ & IQA $\uparrow$ & ASE $\uparrow$ & Sync-C $\uparrow$ & Sync-D $\downarrow$ \\
    \hline
   Sonic & 46.47/87.61 & 213.15/232.65 & 7.53/6.37 & 4.58/3.11 & 6.91/5.28 & 8.57/8.15 \\
   Hallo3 & 33.16/80.17 & 185.40/159.04 & 7.96/7.15 & 4.81/3.76 & 6.55/4.64 & 9.01/9.17 \\
   FantasyTalking & 38.17/78.72 & 86.89/\underline{138.22} & 7.62/7.16 & 4.83/3.82 & 3.56/3.22 & 11.16/10.14 \\   
   HunyuanVideo-Avatar & 34.80/78.85 & 175.00/230.41 & 7.95/7.29 & 5.13/4.06 & 7.43/4.81 & 8.12/8.11 \\
   MultiTalk & 38.51/\underline{77.92} & 172.02/206.46 & \underline{8.35}/7.24 & \underline{5.71}/3.95 & \textbf{8.57}/\textbf{5.64} & \textbf{6.97}/\underline{7.67} \\
   OmniAvatar & 36.19/82.40 & 137.19/169.66 & 8.14/\textbf{7.35} & 5.35/\textbf{4.14} & 7.72/5.36 & 7.66/7.76 \\
   \hline 
   Ours (w/o DPO) & \underline{29.05}/\underline{76.25} & \underline{86.10}/152.33 & 7.94/7.27 & 5.66/3.99 & 7.89/5.28 & 7.53/7.84 \\ 
   Ours (w/ DPO) & \textbf{27.63}/\textbf{66.11} & \textbf{81.86}/\textbf{133.78} & \textbf{8.38}/\underline{7.33} & \textbf{5.96}/\underline{4.13} & \underline{8.15}/\underline{5.49} & \underline{7.32}/\textbf{7.66} \\
  \bottomrule
  \end{tabular}%
  \caption{Quantitative comparisons of video quality and lip synchronization with other competing methods on two test datasets. The best results are in bold, and the second-best are underlined.}
  \label{tab:quantitative_comparison}%
\end{table*}

\begin{figure*}[!t]
\centering
\includegraphics[width=0.95\textwidth]{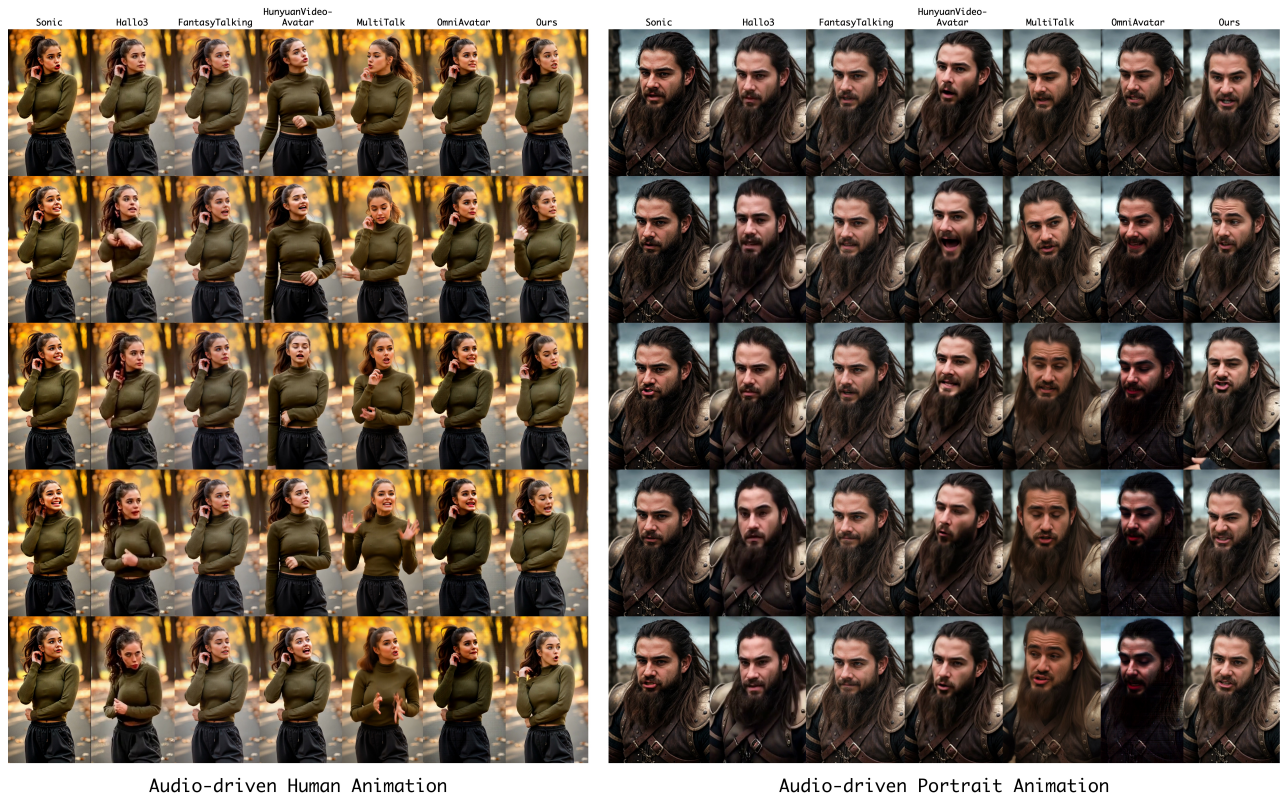}
\caption{Qualitative comparison with other competing methods.}
\label{fig:qualitative_res}
\end{figure*}

\subsection{Results and Analysis}
\label{subsec:4_2}
We conduct both qualitative and quantitative evaluations of our method by comparing it with SOTA audio-driven animation approaches, including Sonic, Hallo3, FantasyTalking, HunyuanVideo-Avatar, MultiTalk, and OmniAvatar. Since the work Hallo4 has not yet released its code and models, a direct comparison is not feasible.

\subsubsection{Quantitative Results.}
As shown in \cref{tab:quantitative_comparison}, our method significantly outperforms existing approaches in FID and FVD across both test datasets. On the HDTF benchmark, we achieve the best results in all image and video quality metrics (FID, FVD, IQA, ASE) and performs competitively in lip synchronization. On the CelebV-HQ test set, our method achieves the best scores in FID, FVD, and Sync-D, and ranks second in the remaining metrics(IQA, ASE, and Sync-C), with only a marginal gap to the best result. Overall, our method delivers superior quantitative performance compared to current SOTA methods.

\subsubsection{Qualitative Results.}
We conducted qualitative comparisons with existing methods. As shown in \cref{fig:qualitative_res}, for human animation, our method generates videos with more natural variations in the foreground, background, and character movements, as well as higher overall quality. In contrast, Sonic, HunyuanVideo-Avatar, and OmniAvatar produce unnatural facial expressions and inaccurate lip synchronization, while FantasyTalking exhibits motion only in the mouth region, with minimal changes elsewhere. Hallo3 and MultiTalk show noticeable artifacts in the face and hands. For portrait animation, Hallo3, HunyuanVideo-Avatar, and MultiTalk fail to maintain character consistency, whereas FantasyTalking animates only the mouth with limited motion in other areas. Sonic demonstrates limited facial expressiveness, and OmniAvatar suffers from severe color distortion. In comparison, our method generates more natural and vivid facial expressions and more aesthetically pleasing visual effects, resulting in superior video quality.

\begin{figure}[t]
\centering
\includegraphics[width=0.95\columnwidth]{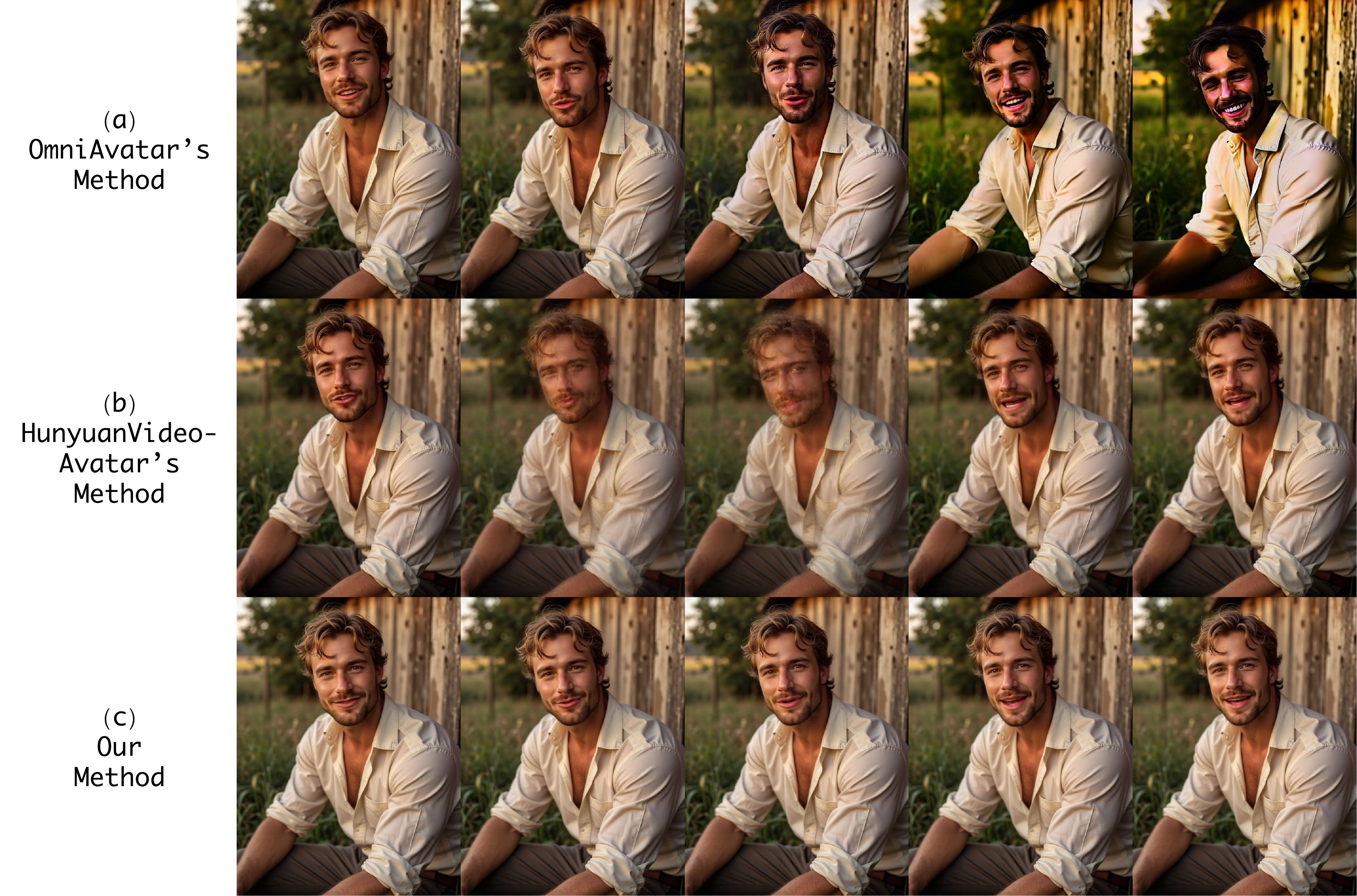}
\caption{Qualitative comparison of long video generation results.}
\label{fig:ab_lvg}
\end{figure}

\begin{table}[!h]
  \centering
  \renewcommand{\arraystretch}{1.3}
  \begin{tabular}{lcccc}
  \toprule
  Methods & LS $\uparrow$ & VD $\uparrow$ & N $\uparrow$ & VA $\uparrow$ \\
  \hline
  Sonic & 3.50 & 3.14 & 3.21 & 3.21 \\
  Hallo3 & 2.86 & 2.79 & 2.79 & 2.86 \\
  FantasyTalking & 1.93 & 2.64 & 2.57 & 2.71 \\
  HunyuanVideo-Avatar & 3.54 & 3.14 & 3.05 & 2.86 \\
  MultiTalk & \underline{3.93} & \underline{3.79} & \textbf{3.93} & \underline{3.79} \\
  OmniAvatar & 3.71 & 3.77 & 3.21 & 3.29 \\
  \hline
  Ours & \textbf{4.02} & \textbf{3.98} & \underline{3.90} & \textbf{4.11} \\
  \bottomrule
  \end{tabular}
  \caption{User Study results. The best results are in bold, and the second-best are in underlined.}
  \label{tab:user_study}
\end{table}

\subsubsection{User Study.}
To further validate the effectiveness of our proposed method, we conducted a user study with 50 participants, who rated the videos using a 5-point Mean Opinion Score (MOS) scale across four critical dimensions: Lip Synchronization (LS), Video Definition (VD), Naturalness (N), and Visual Appeal (VA). As shown in \cref{tab:user_study}, our method achieves higher scores in LS, VD, and VA. Although the naturalness score is slightly lower than that of MultiTalk, it still significantly outperforms all other methods, demonstrating competitive performance. This comprehensive evaluation highlights the superiority of our approach in generating realistic and diverse talking animations while maintaining consistent identity representation and high visual fidelity.

\subsection{Ablation Studies}
\label{subsec:4_3}
\subsubsection{Ablation on Long Video Generation.}
We conducted ablation experiments on the LoRA-based long video generation method described in Section 3.1. Specifically, we trained a model without incorporating the improvements outlined in that section, and then generated long videos using the approaches from OmniAvatar and HunyuanVideo-Avatar. As illustrated in \cref{fig:ab_lvg}, the final latent extension strategy used in OmniAvatar suffers from error accumulation over time, leading to significant degradation in the quality of the generated long video. The Time-aware Position Shift Fusion method employed in HunyuanVideo-Avatar produces visible artifacts in the transition regions due to the special input format of the DiT backbone. In contrast, our method generates temporally coherent and identity-consistent long videos, effectively preserving both visual fidelity and temporal smoothness.

\begin{figure}[t]
\centering
\includegraphics[width=0.95\columnwidth]{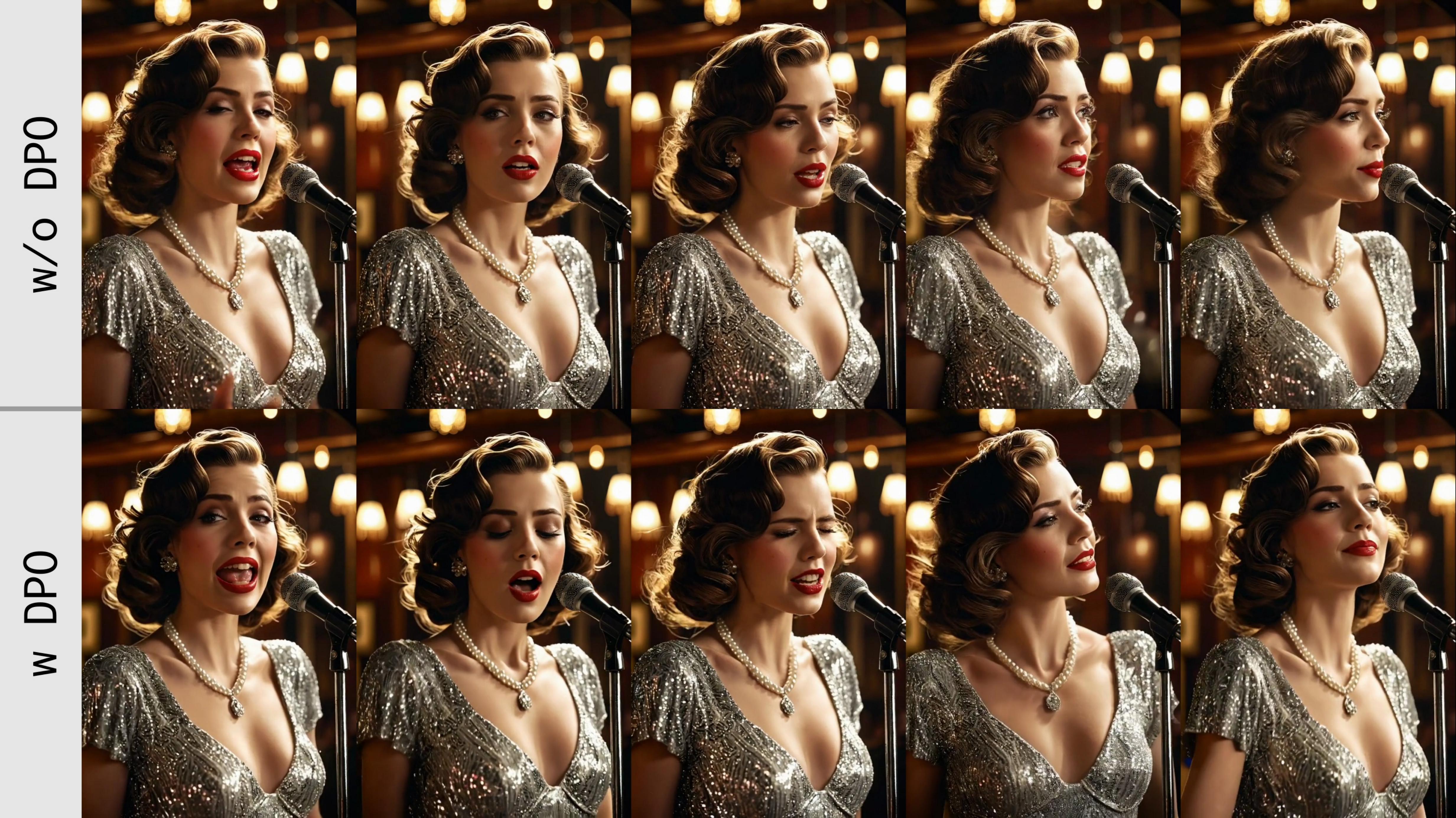}
\caption{Visual comparison of DPO ablation study.}
\label{fig:ab_dpo}
\end{figure}

% \begin{table*}[!t]
%     \centering
%     \renewcommand{\arraystretch}{1.4}
%     \begin{tabular}{lcccccc}
%     \toprule
%     \multirow{2}{*}{Method} & \multicolumn{6}{c}{HDTF/CelebV-HQ} \\
%     % \cline{2-7}
%     ~ & FID $\downarrow$ & FVD $\downarrow$ & IQA $\uparrow$ & ASE $\uparrow$ & Sync-C $\uparrow$ & Sync-D $\downarrow$ \\
%     \hline
%     w/o DPO & 29.05/76.25 & 86.10/152.33 & 7.94/7.27 & 5.66/3.99 & 7.89/5.28 & 7.53/7.84 \\
%     w/ DPO & \textbf{27.63}/\textbf{66.11} & \textbf{81.86}/\textbf{133.78} & \textbf{8.38}/\textbf{7.33} & \textbf{5.96}/\textbf{4.13} & \textbf{8.15}/\textbf{5.49} & \textbf{7.32}/\textbf{7.66}  \\
%     \bottomrule
%     \end{tabular}
%     \caption{Ablation study of the reward feedback.}
%     \label{tab:ab_dpo}
% \end{table*}

\subsubsection{Ablation on the Reward Feedback.}
We train models with and without DPO to evaluate the Reward Feedback method (\cref{subsec:3_2}) both quantitatively and qualitatively. As shown in \cref{tab:quantitative_comparison}, incorporating DPO leads to consistent improvements across all metrics, indicating enhanced video quality and lip synchronization accuracy. Qualitatively (\cref{fig:ab_dpo}), the model with DPO generates rich, context-appropriate facial expressions for singing audio, while the ablated version produces flat and under-expressive results. These results demonstrate that our DPO-based approach improves not only fidelity and synchronization but also expressiveness, yielding outputs better aligned with human preferences.

\section{Conclusion}
We present a novel DiT-based framework for high-quality, audio-driven human video generation, addressing key challenges in long-sequence temporal coherence and multi-character animation. Our method enables arbitrarily long video generation via LoRA-based training and a position shift inference technique, preserving temporal coherence, identity consistency, and the integrity of the pre-trained model. To further enhance synchronization and motion naturalness, we introduce a partial parameter update scheme combined with reward feedback, which improves both lip synchronization accuracy and upper-body dynamics. Furthermore, we propose Mask-CFG, a training-free approach for multi-character animation that requires no additional data or model fine-tuning, yet supports audio-driven animation of three or more characters. To the best of our knowledge, this is the first training-free method to enable audio-driven animation for three or more characters. Extensive experiments show that our method surpasses existing SOTA approaches in terms of visual quality, temporal consistency, and scalability.

\bibliography{aaai2026}

\end{document}